\documentclass[letterpaper, 10 pt, conference]{ieeeconf}  

\IEEEoverridecommandlockouts                              

\usepackage{graphicx} 
\usepackage{amsmath} 
\usepackage{amssymb}  
\usepackage{siunitx}
\usepackage{url}

\title{\LARGE \bf
Teaching Robots Novel Objects by Pointing at Them
}

\author{Sagar Gubbi Venkatesh$^{1, 2}$, Raviteja Upadrashta$^{2}$, Shishir Kolathaya$^{2}$, and Bharadwaj Amrutur$^{1, 2}$
\thanks{*This work was supported by Robert Bosch Center for Cyber-Physical Systems.}
\thanks{$^{1}$Department of Electrical and Communication Engineering, Indian Institute of Science, Bangalore 560012, India {\tt\small sagar@iisc.ac.in}; {\tt\small amrutur@iisc.ac.in}}%
\thanks{$^{2}$Robert Bosch Center for Cyber-Physical Systems, Indian Institute of Science, Bangalore 560012, India {\tt\small ravitejaupadras@iisc.ac.in}; {\tt\small shishirk@iisc.ac.in}}%
}

\begin{document}

\maketitle
\thispagestyle{empty}
\pagestyle{empty}

\begin{abstract}
Robots that must operate in novel environments and collaborate with humans must be capable of acquiring new knowledge from human experts during operation. We propose teaching a robot novel objects it has not encountered before by pointing a hand at the new object of interest. An end-to-end neural network is used to attend to the novel object of interest indicated by the pointing hand and then to localize the object in new scenes. In order to attend to the novel object indicated by the pointing hand, we propose a spatial attention modulation mechanism that learns to focus on the highlighted object while ignoring the other objects in the scene. We show that a robot arm can manipulate novel objects that are highlighted by pointing a hand at them. We also evaluate the performance of the proposed architecture on a synthetic dataset constructed using emojis and on a real-world dataset of common objects.

\end{abstract}

\section{INTRODUCTION}

\begin{figure}[!t]
    \centering
    \includegraphics[width=1.0\linewidth]{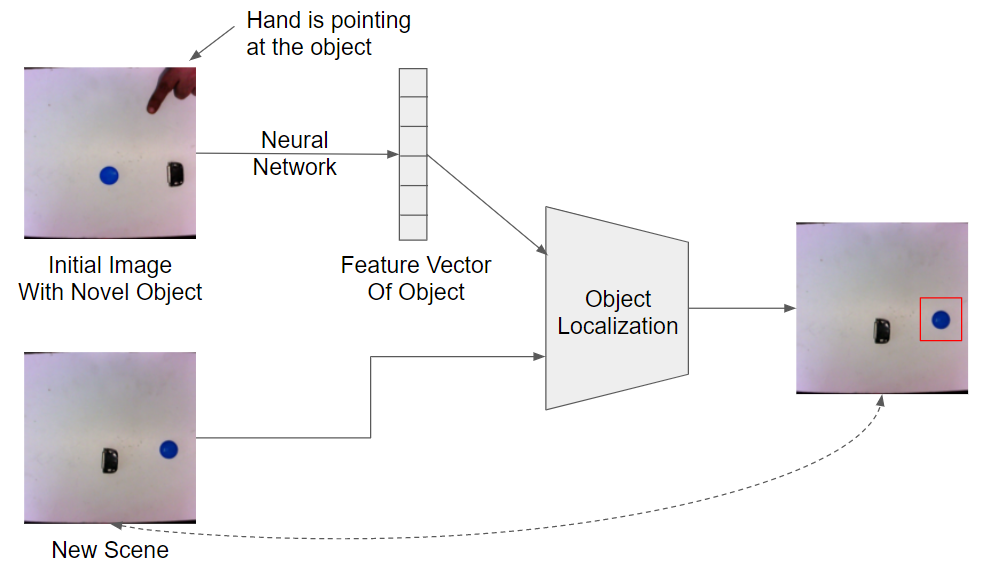}
    \caption{One-shot localization of novel object selected by the pointing hand. The feature vector of the object (blue bottle cap) that the hand is pointing at is extracted using the proposed attention modulation mechanism. This is then used to localize the object in new scenes. Note that the pointing finger is at considerable distance from the object of interest.}
    \label{fig:hero}
\end{figure}

Robots that can operate in unconstrained environments and collaborate with humans must be capable of learning about new objects they may encounter. Pointing at an object with our hand is a natural way to communicate with the robot about a new object. In this paper, we consider the problem of teaching robots novel objects by pointing at the new object. Once we show the robot a new object, it can generate and store a feature vector corresponding to that object and then re-use it for one-shot localization of the object in new scenes (Fig.~\ref{fig:hero}).

Neural networks have been used in recent years to learn fully differentiable visuomotor policies that directly map pixels to actuator commands~\cite{levine2016end,zhang2018deep,yu2018one,rahmatizadeh2018virtual,giusti2015machine}. The neural network architecture typically used for such policies can be decomposed into vision layers and control layers. The vision layers comprise of several convolutional and pooling layers followed by a spatial attention mechanism that attends to the objects of interest in the image. We propose modulating the spatial attention so as the network is able to attend to the object that the hand is pointing at (see Fig.~\ref{fig:hero}) while ignoring the other distracting objects in the scene.

In this work, we assume that only the location of the object of interest is available for training and that the position and orientation of the pointing hand are unavailable. So, this is a weakly supervised learning problem where the neural network must figure out as part of the learning process that the pointing hand in the image is salient and then learn to attend to the object being pointed at. On the other hand, this assumption makes the process of data acquisition with a real robot easier by reducing the labeling effort.

Unlike other papers on object detection~\cite{redmon2016you,ren2015faster}, we are primarily interested in teaching robots new objects. This means that we are interested in objects not seen by the neural network during training. We accomplish this using Siamese networks~\cite{koch2015siamese,venkatesh2019one,bertinetto2016fully}, which are twin neural networks with shared weights. The idea is to use the neural network to obtain from the image a feature vector representing the object of interest rather than classifying the contents of the image as is usually done (Fig.~\ref{fig:hero}). This vector can be subsequently used in new environments to find the novel object of interest.

Our contributions are as follows:
\begin{itemize}
    \item We propose a spatial attention modulation mechanism that endows the neural network with the ability to selectively attend to the object that is being pointed at while ignoring other distracting objects in the scene.
    \item We show that the proposed method can be combined with Siamese networks to teach robots novel objects.
\end{itemize}

The proposed network architecture is trained on synthetic data constructed from a dataset of emojis. We demonstrate the proposed method on the Dobot Magician robot arm. We show that the robot learns new objects that we point at and can find them in new scenes.

The rest of this paper is organized as follows. In the next section, papers related to this work are discussed. In Section~3, the proposed model architecture is described. Section~4 details experimental results, and Section~5 concludes the paper.

\section{RELATED WORK}
\subsection{Hand Recognition}
One of the ways to design a system that can infer the object of interest in a similar scenario is to use a pipelined approach. For example, one could employ deep learning models trained to localize human hands in an image~\cite{openpose, doosti2019hand}, extract the most relevant keypoints of the hand (say the joints of the index finger) and then fit a line that passes through these keypoints. An object recognition module could then be used to localize each object in the scene, project these points on the line and pick the object corresponding to the closest point to the the hand as the object of interest. However, training such a system requires a strong level of supervision such as the positions of all objects in the scene, and possibly even the keypoints of the hand if one would like to fine-tune the hand localization models. However, this approach will not be feasible in a weak supervision setting as outlined in this paper where only the location of the object of interest is given. Moreover, it has been shown across a wide range of problems that using an end-to-end approach leads to better performance as compared to using a pipelined approach for a given task~\cite{gupta2016synthetic, end_to_end_speech, zhang2018deep}.

\subsection{Spatial Attention}
The architecture of typical end-to-end networks for visuomotor tasks can be broadly grouped into two sets of layers. The initial group of layers form the vision layers that help in localizing the relevant objects in the image. The remaining layers form what is known as the control layers which are responsible for coming up with the appropriate control actions required to perform the task at hand. A key component in such end-to-end networks is some form of a spatial attention mechanism that learns to attend to the relevant object of interest in the scene. The work presented in~\cite{zhang2018deep} demonstrates the use of imitation learning for teaching a PR2 robot to perform simple tasks such as pick-and-place. The authors developed a virtual reality based system to teleoperate the robot and collect training data. The data was then used to train an end-to-end network that maps image pixels directly to robot joint velocities. The network consists of an initial set of convolution layers that generates a feature map. The feature map is passed through a spatial softargmax layer to output a feature vector. The resulting feature vector is then passed through a few fully connected layers to predict the joint velocities of the robot. The spatial softargmax layer serves as a simple spatial attention mechanism where the attention weight corresponding to each pixel of the feature map depends on the degree of activation.

\subsection{One Shot Learning}
Apart from inferring the object of interest in an image in the presence of other objects, another goal of this paper is to enable robots to recognize objects that they have never encountered before by training it on only a few examples involving the novel object. Broadly speaking, meta learning and Siamese networks are two approaches one can take to achieve this. We review both approaches below.

\subsubsection{Meta Learning}
In Meta learning, also known as learning to learn, a distribution of tasks are provided as training data. Typically only a few examples for each task are provided in the training data. The weights of the network after the training process completes serves as a good initialization for the network to learn to perform any new previously unseen task. Only a few training examples involving the novel task and a few gradient descent steps are required for the network to converge to an optimal set of weights~\cite{maml}. Recently, meta learning and imitation learning has been combined to enable robots to perform novel tasks such as pick-and-place by training on just a single example \cite{osiml, daml}.

\begin{figure*}[!t]
    \centering
    \includegraphics[width=0.99\linewidth]{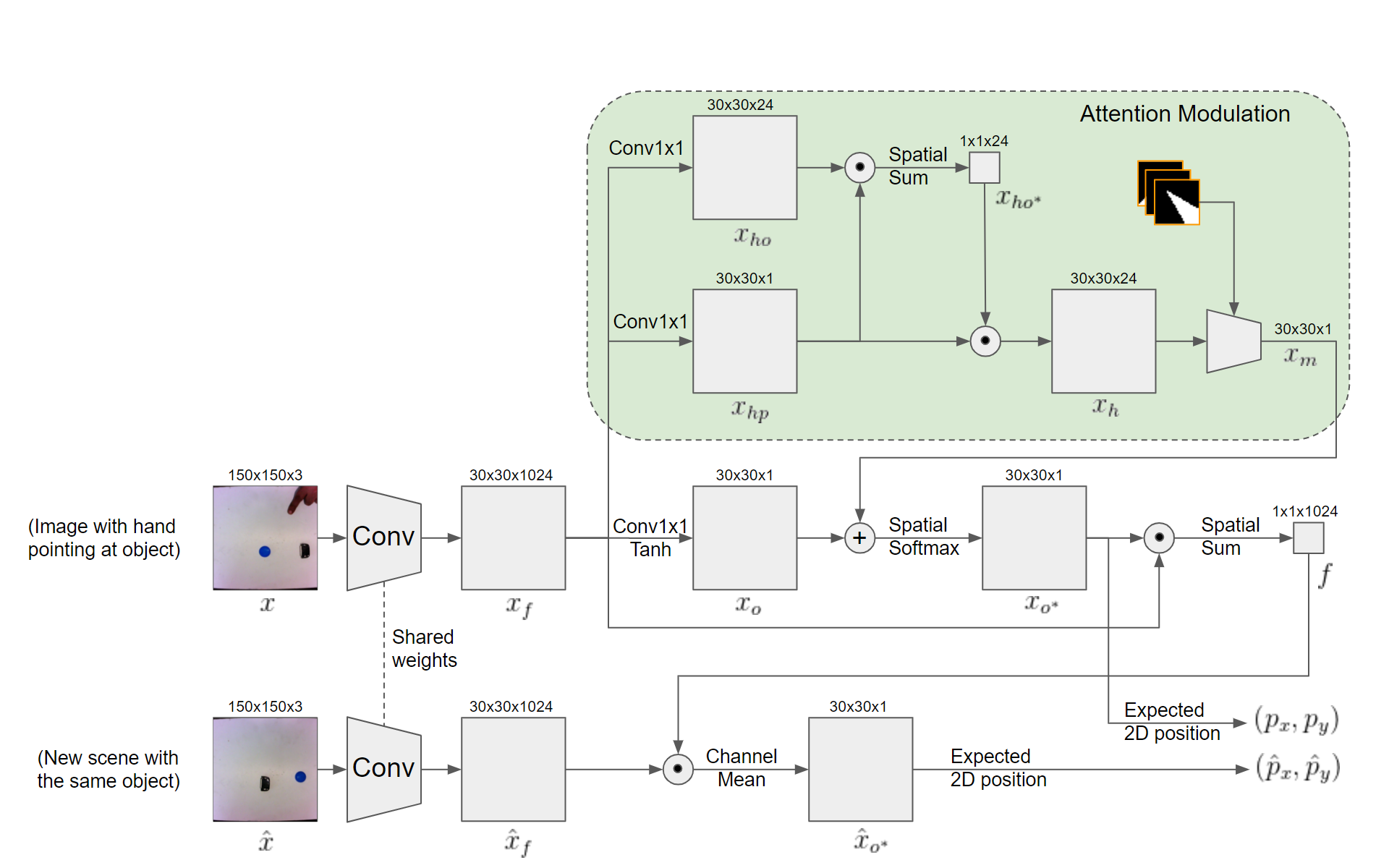}
    \caption{The proposed neural network architecture for one-shot localization of the object selected by the pointing hand. The convolutional layers in the ``Conv" block are Conv3x3(16)-ELU-Conv3x3(32)-ELU-Conv3x3(64)-ELU-MaxPool2x2-Conv3x3(64)-ELU-Conv3x3(128)-ELU-MaxPool2x2-Conv3x3(256)-ELU-Conv3x3(512)-ELU-Conv3x3(1024)-ELU. All convolutions are ``valid" convolutions that do not use padding so that the feature vector for the object is the same regardless of whether it is near the edge of the image or at the center. The receptive field of the ``Conv" block is 34~px.}
    \label{fig:arch}
\end{figure*}

\subsubsection{Siamese Networks}
Siamese networks are used to address the similarity learning problem where it is desirable to infer if a pair of images (referred to exemplar and search images) are similar to each other or not. This is done by using twin convolution neural networks with shared weights that transform the images $x_1$ and $x_2$ into feature embeddings $\phi(x_1)$ and $\phi(x_2)$, respectively. The embedding pair is then combined using a transformation $g$ that can be used to make suitable predictions depending on the task at hand. For example, in the context of image classification~\cite{koch2015siamese} the transformation $g$  is a distance metric that can be used to measure the similarity score between the object in the exemplar image and the search image. The system is trained on several examples of similar and dissimilar pairs of images. Once training is complete a database of images is built with one image corresponding to each object of interest. At test time the similarity of the search image is tested against each image in the database to determine the object class of the search image. Siamese networks have been used in face recognition systems as well~\cite{deepface}. However, in both these papers the comparison is possible only if the exemplar and search images are of the same dimension. 

The authors of~\cite{bertinetto2016fully} used fully convolutional neural networks to enable comparison of images of dissimilar dimensions with Siamese networks. Their architecture was adapted successfully for object tracking in videos. Here the user provides the exemplar image by cropping out the object of interest from the first frame of the video which is then compared against each subsequent frame using the Siamese network. More recently, the authors in~\cite{venkatesh2019one} combined fully convolutional Siamese networks with spatial attention to enable object localization for robot pick-and-place tasks. The paper explores specifying the object of interest by using visual cues instead of requiring the user to provide a cropped image of the object. Given a group of objects in a scene the user indicates to the robot the object of interest by shining a laser beam directly on it. Although the authors talk of localizing novel objects using a laser beam as a visual cue, the network designed by them should work for any other kind of visual cue (such as a stick or even a hand) so long as it is in very close proximity with the object of interest. However, a human merely has to point at an object from a distance to convey that it is of interest and an observer infers and localizes the object being referred to by looking in the direction of the pointing hand. We would like to design systems that communicate intent to robots much like how humans communicate with each other via visual cues or using natural language~\cite{blocks, touchdown}. Learning to localize an object of interest from natural language instructions requires a different architecture design compared to the one presented in this paper. We will restrict our focus to learning to localize the object of interest by using a visual cue provided such as a hand pointing at the object from a distance.

\section{NETWORK ARCHITECTURE}

\subsection{Localizing the Object of Interest}

The proposed neural network for one-shot localization is shown in Fig.~\ref{fig:arch}. Let the exemplar image (denoted as $x$) correspond to the image that contains the novel object that is being pointed at by the hand. Let the search image (denoted as $\hat{x}$) be the image of the new scene in which the same object must be localized. The network outputs the locations of the object in the exemplar image and the search image which are denoted as $(p_x, p_y)$ and $(\hat{p}_x, \hat{p}_y)$, respectively. The mean squared error loss is used to train the network.

The localization of the object is performed in a similar fashion as described in \cite{venkatesh2019one} except for the attention modulation block. The exemplar image is passed through the CNN to obtain a feature map $x_f$. This is then passed through a bottleneck convolutional layer (conv$1\times1$) to obtain $x_o$. Let us ignore for now how the attention modulation map $x_m$ is generated. We will describe the generation of the attention map $x_m$ in Section~\ref{sec:attn-mod}. A spatial attention map $x_{o^*}$ is generated  by adding the attention modulation map $x_m$ to $x_o$ and the resulting sum is passed through a spatial soft-argmax layer whose output is the predicted location of the object of interest $(p_x, p_y)$ in the exemplar image $x$ (see Eqns. \eqref{eqn:x_ostart}, \eqref{eqn:softargmax_x} and \eqref{eqn:softargmax_y}). The spatial attention map $x_{o^*}$ is used to obtain the feature vector $f$ corresponding to the object of interest from the feature map $x_f$ (see Eqn. \eqref{eqn:feat_vec}). Note that $\odot$ in Fig.~\ref{fig:arch} corresponds to the element-wise multiplication operation (with the appropriate broadcasting done to account for the different number of channels present in $x_f$ and $x_{o^*}$).


\begin{equation}
    x_{o^*_{i, j}} = softmax_{i, j} \big(x_{o_{i, j}} + x_{m_{i, j}}\big)
    \label{eqn:x_ostart}
\end{equation}

\begin{equation}
    p_x = \sum_{i, j} x_{o^*_{i, j}} i
    \label{eqn:softargmax_x}
\end{equation}
\begin{equation}
    p_y = \sum_{i, j} x_{o^*_{i, j}} j
    \label{eqn:softargmax_y}
\end{equation}

\begin{equation}
    f = \sum_{i, j} x_{o^*_{i, j}} x_{f_{i, j}}
    \label{eqn:feat_vec}
\end{equation}

The localization of the object in the search image $\hat{x}$ is done by first passing $\hat{x}$ through the CNN to obtain $\hat{x}_f$. Then the feature vector $f$ is used like a matched filter (or equivalently as a conv1$\times$1 layer with $f$ as the weights) to generate $\hat{x}_{o^*}$. The location of the object $(\hat{p}_x,\hat{p}_y)$ in the search image $\hat{x}$ is then determined by passing $\hat{x}_{o^*}$ through a spatial soft-argmax layer (similar to the operations in Eqns.~\eqref{eqn:softargmax_x} and \eqref{eqn:softargmax_y}).

\subsection{Generating the Attention Modulation Map}
\label{sec:attn-mod}

\begin{figure}[!t]
    \centering
    \includegraphics[width=1.0\linewidth]{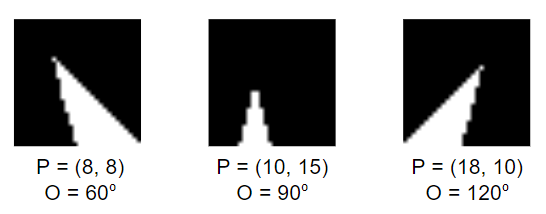}
    \caption{Beam / Cone like attention modulation maps for different positions and orientations of the pointing hand. The dark regions correspond to a value of -2.0 which suppresses peaks corresponding to objects in $x_o$ (Fig.~\ref{fig:arch}), whereas the bright regions correspond to value 0.0 which allows values in that area in $x_o$ to pass through unchanged. The beam width is \ang{30}, and the step size is \ang{15}.}
    \label{fig:attn_mod}
\end{figure}

When there are multiple objects in the scene, we expect multiple bright spots each corresponding to an object in $x_o$ (Fig.~\ref{fig:sample_non_siamese}). We would like to suppress the peaks in $x_o$ corresponding to objects that are not being pointed at. Had we known the location and orientation of the hand, we could have directly suppressed the irrelevant peaks. However, since we do not have labels corresponding to the pose of the pointing hand in the scene, the neural network must learn to attend to the hand and then use this to  suppress the irrelevant peaks in $x_o$. To enable this, we use a ``soft" or differentiable way to compute the position and orientation of the hand which is then employed to suppress irrelevant objects in $x_o$.

The feature map $x_f$ is passed through two independent bottleneck layers to produce maps $x_{hp}$ and $x_{ho}$ corresponding to the position and orientation of the pointing hand respectively in $x$. Similar to Eq.~\eqref{eqn:feat_vec}, spatial attention is used to attend to the pointing hand and to obtain the orientation of the hand $x_{ho^*}$. The final position and orientation of the hand ($x_h$) is used to ``soft select" a pre-defined attention modulation map $x_m$ (see Fig.\ref{fig:attn_mod}). The set of pre-defined attention modulation maps include beams from all possible locations and orientations of the pointing hand as shown in Fig.~\ref{fig:attn_mod}. Each modulation map is constructed by drawing a beam emanating from the position of the hand and in the direction the hand is pointing at. We use an orientation step size of $\ang{15}$ with a beam width of $\ang{30}$. Thus, there are 30$\times$30$\times$24 $=$ 21600 such maps. Note that no explicit loss function is used to learn $x_{hp}$ and $x_{ho}$. Rather, network learns to predict appropriate values for $x_{hp}$ and $x_{ho}$ that result in the ``selection" of an appropriate attention map, which is possible only by correctly recognizing the position and orientation of the hand. The modulation map thus obtained, $x_m$, is added to $x_o$ to highlight the object being pointed at while suppressing the irrelevant ones. Thus, the pixels in $x_o$ that lie inside the beam are passed as is whereas the pixels that lie outside the beam are suppressed. Note that the entire attention modulation scheme is differentiable and hence can be learned through back-propagation.

The proposed way of creating attention modulation maps is most suitable for top view images (Fig.~\ref{fig:sample_non_siamese}). For perspective views where the depth of the object is more relevant, it may be necessary to generate maps in 3D by casting a cone of rays and using the perspective projection. We leave this for future work.



\section{EXPERIMENTAL RESULTS}

To evaluate the proposed neural network, we first train it on a synthetic dataset and compare it with alternative architectures. The trained network is deployed on a robot arm to demonstrate its real world performance.

\subsection{Localization Performance}
\begin{figure}[!t]
      \centering
      \includegraphics[width=0.99\linewidth]{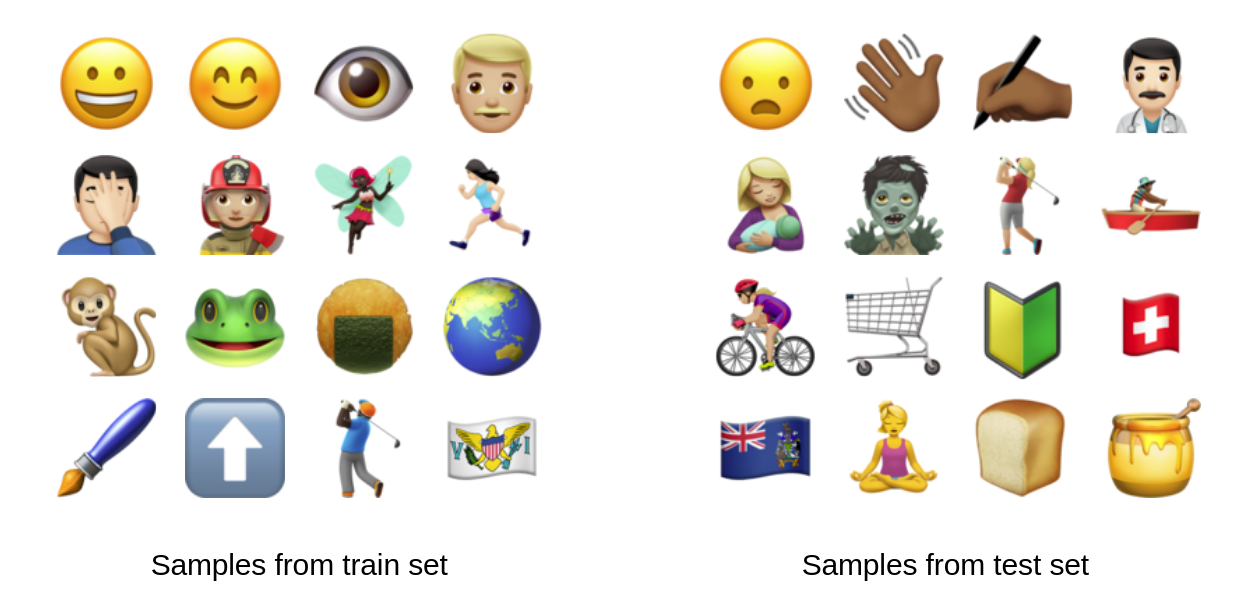}
      \caption{A few sample objects used for training and evaluation. The set of emojis is divided into 2075 for training and 703 for testing.}
      \label{fig:emoji}
\end{figure}

\begin{figure}[!t]
      \centering
      \includegraphics[width=0.9\linewidth]{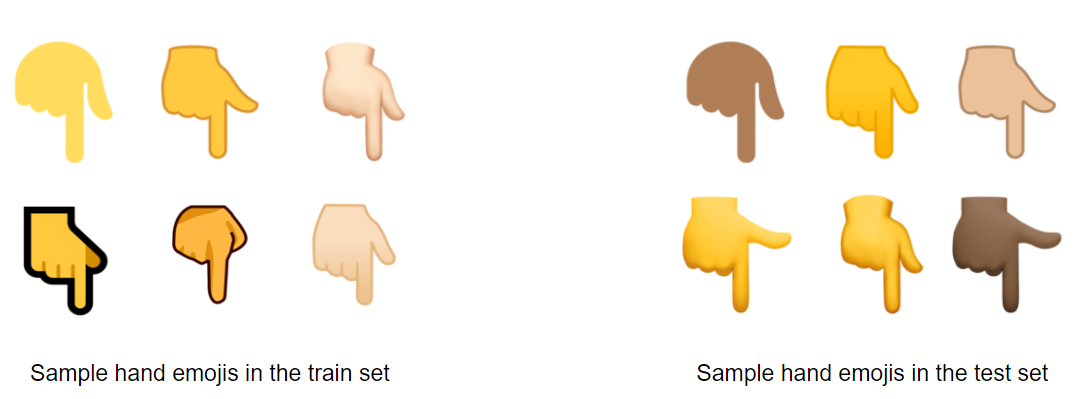}
      \caption{A few sample hand images used for training and evaluation. The set of hand emojis is divided into 47 for training and 8 for testing.}
      \label{fig:hand_emoji}
\end{figure}

\begin{figure}[!t]
      \centering
      \includegraphics[width=0.99\linewidth]{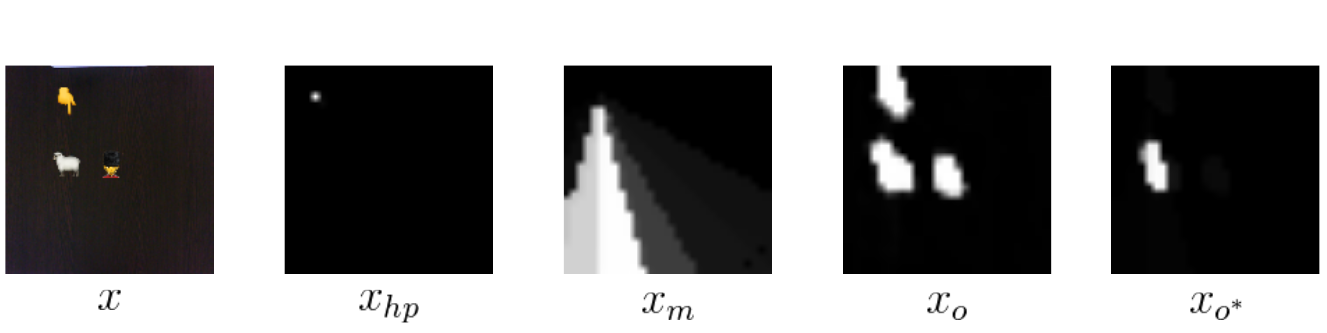}
      \caption{Sample prediction of the proposed architecture. The network has properly localized the pointing hand and chosen a suitable attention modulation map. The activation corresponding to the object that is not pointed at has been appropriately suppressed in $x_{o^*}$.}
      \label{fig:sample_non_siamese}
\end{figure}

A dataset of 5000 training images and 1000 test images is created by placing emojis (Fig.~\ref{fig:emoji}) at non-overlapping positions against a backdrop as shown in Fig.~\ref{fig:sample_non_siamese}. A hand emoji (Fig.~\ref{fig:hand_emoji}) is placed at a random location pointing to an object. One or more distracting objects are placed at random locations not on the line segment between the pointing hand and the object. The label for each sample is the position of the object that the hand is pointing at.

To evaluate the proposed spatial attention modulation mechanism, we first consider only localization of the object in the image containing the pointing hand ($x$) while ignoring the other input ($\hat{x}$) and the output of Siamese network $(\hat{p}_x, \hat{p}_y)$. Table~\ref{table:compare} compares the proposed approach with two baselines. The FC layers baseline refers to using fully connected layers\footnote{The fully connected layers used are FC1024-ELU-FC256-ELU-FC2.} to predict $(p_x, p_y)$ from $x_f$. The Conv layers baseline uses convolutional layers\footnote{Conv3x3(2048)-ELU-MaxPool2x2-Conv3x3(2048)-ELU-Conv3x3(2048)-ELU-MaxPool2x2-Conv3x3(2048)-ELU-Conv3x3(2048)-ELU-FC2.} to predict the position of the object. The networks are trained with mean squared error loss with weight decay 1e-8 using the Adam optimizer\cite{kingma2014adam} with learning rate 1e-4. The evaluation metric is accuracy where we consider the prediction to be accurate if there is sufficient overlap between the ground truth and predicted bounding box. Specifically, the IOU (intersection-over-union) between the ground truth bounding box and the predicted bounding box has to be at least 0.5. All the three networking achieve low training error (accuracy over 99\%), but the test error varies, and the proposed approach generalizes the best. A sample output is shown in Fig.~\ref{fig:sample_non_siamese} where we observe that the spatial attention modulation mechanism is working as one might expect.

\begin{figure}[!t]
      \centering
      \includegraphics[width=0.99\linewidth]{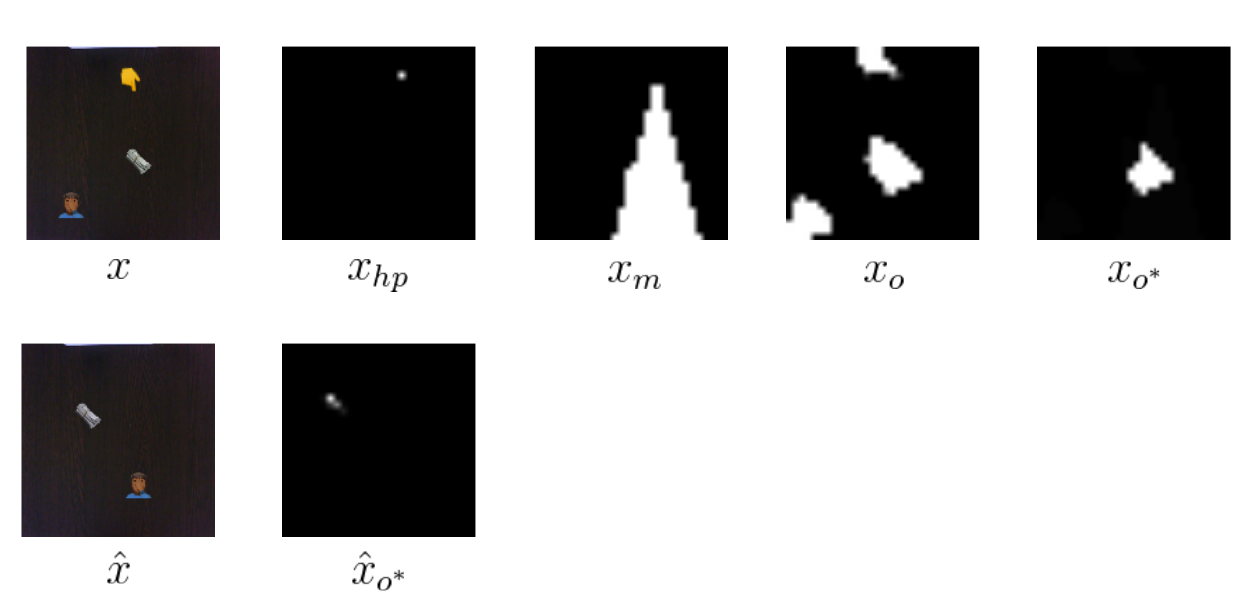}
      \caption{A sample prediction from the proposed architecture on synthetic data.}
      \label{fig:sample_siamese_synth}
\end{figure}

A second dataset containing images corresponding to a new environment ($\hat{x}$) where the object highlighted by the pointing is present along with distracting objects is constructed as before (Fig.~\ref{fig:sample_siamese_synth}). The proposed architecture in its entirety with the Siamese network to process $\hat{x}$ and predict $(\hat{p}_x, \hat{p}_y)$ is trained on this dataset. The accuracy on this dataset drops only marginally to 95.31\%. Table~\ref{table:compare2} compares performance on this dataset and shows that attention modulation is essential to localize the desired object. The sample output in Fig.~\ref{fig:sample_siamese_synth} shows that the desired object in $\hat{x}$ is being attended to.

\begin{table}[!t]
\caption{Comparison of the proposed approach with different baselines}
\label{table:compare}
\begin{center}
\begin{tabular}{ll}
\hline
Neural Network Architecture & Accuracy\\
\hline
FC layers & 11.72\%\\
Conv layers & 41.41\%\\
Proposed approach & 96.88\%\\
\hline
\end{tabular}
\end{center}
\end{table}

\begin{table}[!t]
\caption{Comparison of localization performance of the Siamese network on novel objects with and without attention modulation}
\label{table:compare2}
\begin{center}
\begin{tabular}{ll}
\hline
Neural Network Architecture & Accuracy\\
\hline
Without Attention Modulation\cite{venkatesh2019one} & 12.5\%\\
Proposed approach (with modulation) & 95.31\%\\
\hline
\end{tabular}
\end{center}
\end{table}

\subsection{Evaluation on Robot Arm}
\begin{figure}[!t]
      \centering
      \includegraphics[width=0.85\linewidth]{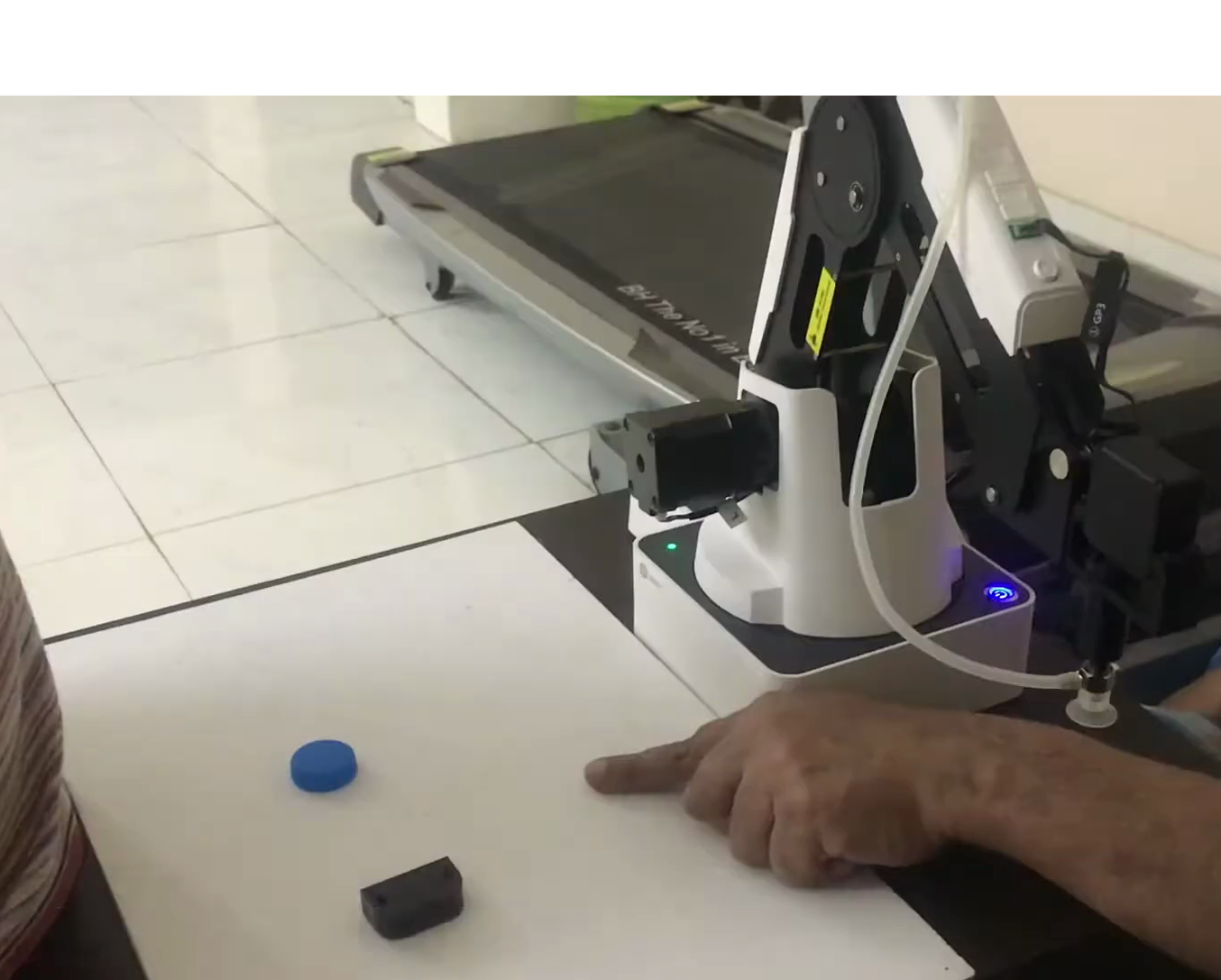}
      \caption{A sample demonstration with the Dobot Magician robot arm.}
      \label{fig:dobot}
\end{figure}

\begin{figure}[!t]
      \centering
      \includegraphics[width=0.85\linewidth]{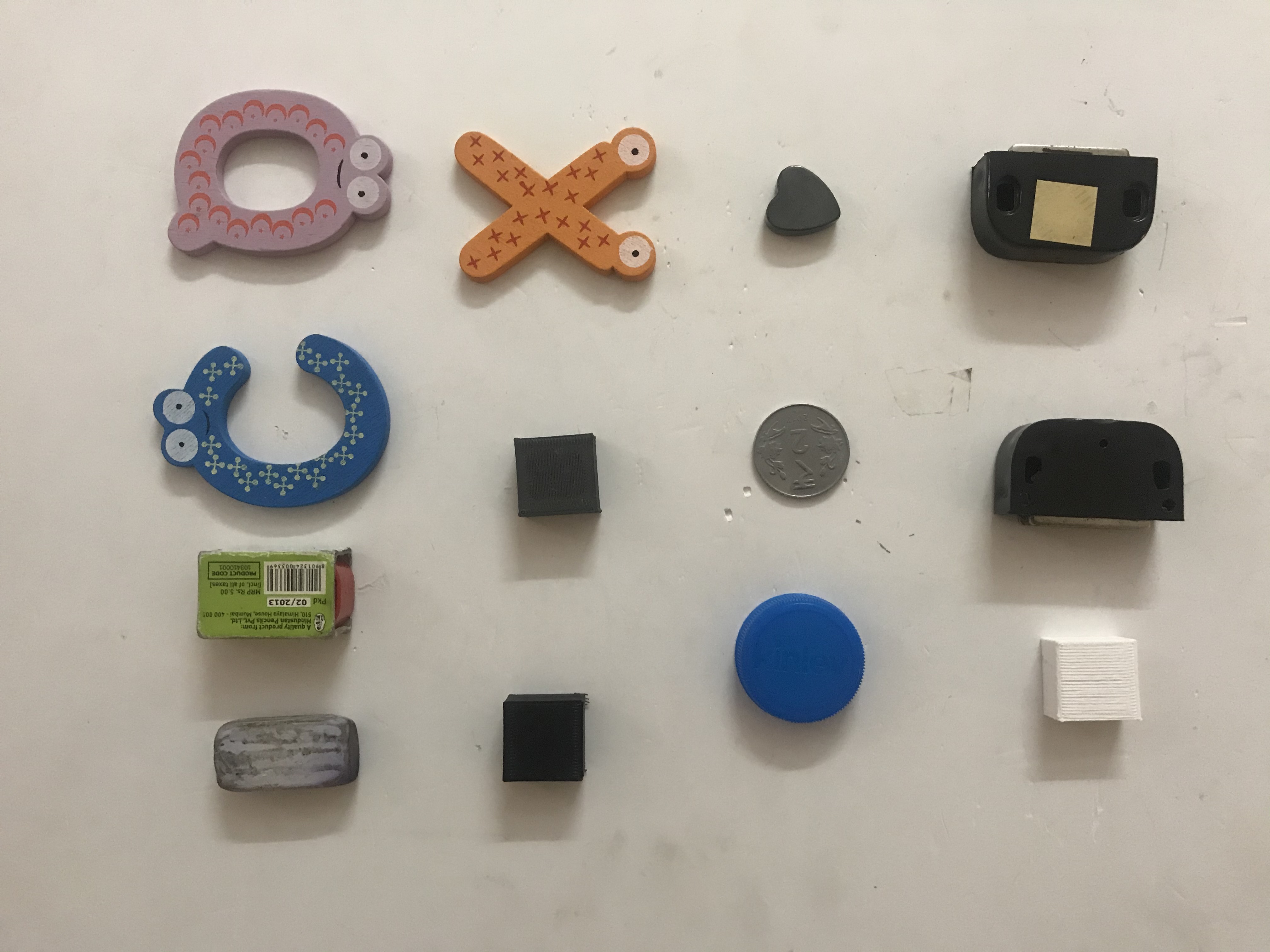}
      \caption{Objects used for evaluating the proposed approach on the Dobot arm.}
      \label{fig:real_objs}
\end{figure}

\begin{figure}[!t]
      \centering
      \includegraphics[width=0.85\linewidth]{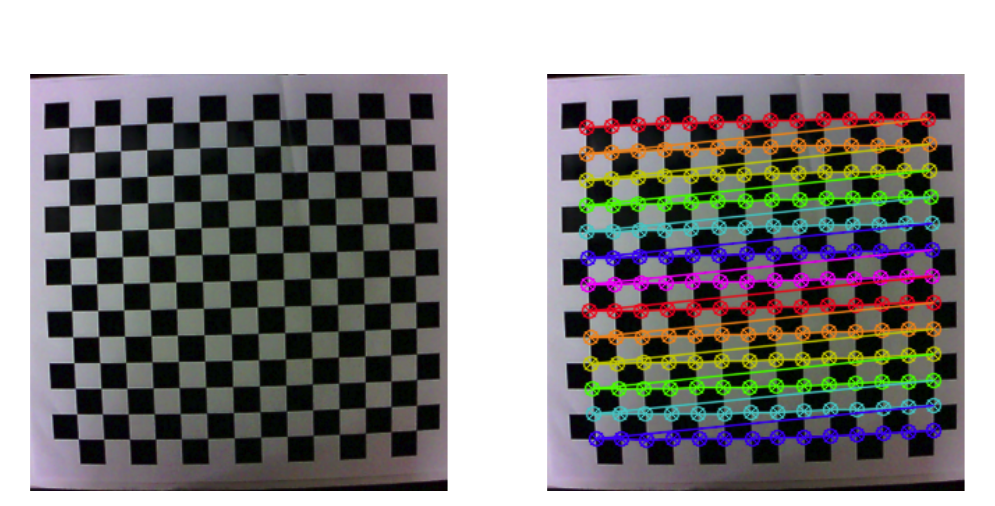}
      \caption{The chessboard calibration pattern used to convert pixels to robot co-ordinates.}
      \label{fig:calib}
\end{figure}

\begin{figure}[!t]
      \centering
      \includegraphics[width=0.99\linewidth]{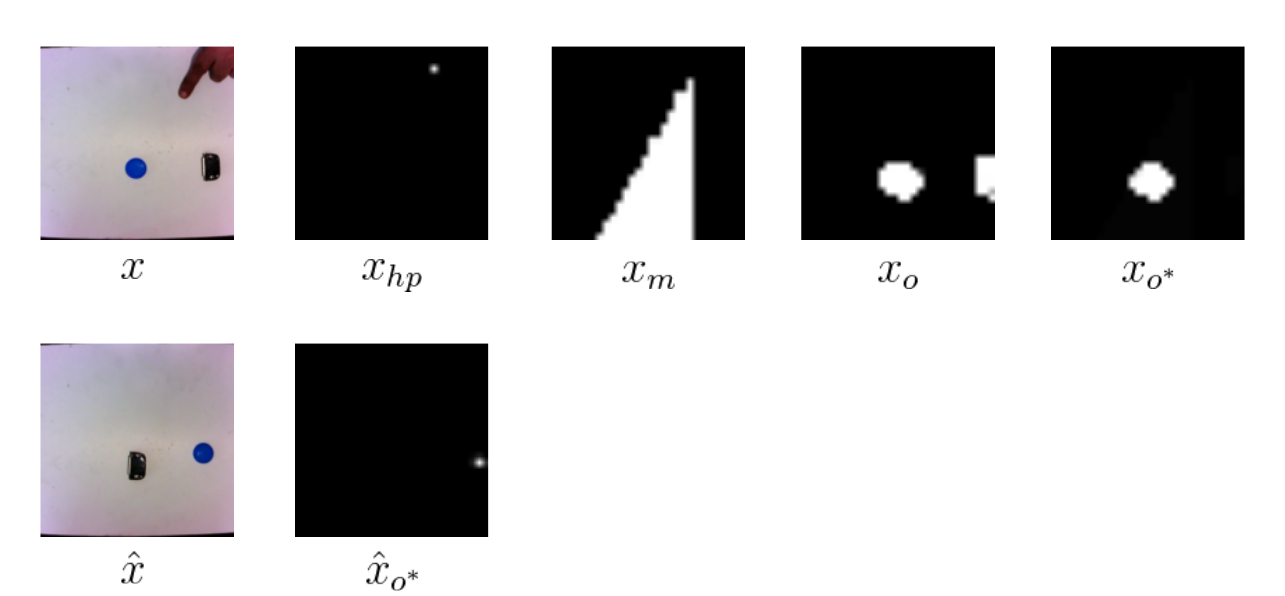}
      \caption{A sample prediction from the proposed architecture.}
      \label{fig:sample_siamese}
\end{figure}

We demonstrate the proposed neural network using the Dobot Magician, a 3-DoF robot arm (Fig.~\ref{fig:dobot}). The objects used for evaluation with the robot are shown in Fig.~\ref{fig:real_objs}. To convert the localized object in pixel space to the robot co-ordinate space, a chessboard calibration pattern is used (Fig.~\ref{fig:calib}), and OpenCV is used for calibration. Figure~\ref{fig:sample_siamese} shows a sample predicted from the proposed neural network. We see that the pointing hand has been localized, and the network has learnt to predict an appropriate attention modulation map that selects the object being pointed at (blue bottle cap in Fig.~\ref{fig:sample_siamese}). We also see that the activation corresponding to the distracting object in $x_{o^*}$ has been successfully suppressed. With the feature vector $f$  corresponding to the bottle cap extracted, the Siamese net successfully attends to the same bottle cap in a new scene ($\hat{x}$). In this manner, 20 trials were performed. The proposed network localizes the desired object to within 1~cm in all the trials. A video of the robot in operation is available at \url{https://youtu.be/bJ5HKllhqLg}.

\section{CONCLUSIONS}

We have proposed a spatial attention modulation method that endows a neural network with the ability to attend to a hand pointing at an object in an image and to focus on the object that is being pointed at. The proposed approach generalizes significantly better compared to architectures that use only fully connected or convolutional layers for localization. Furthermore, this approach can be combined with a Siamese network to localize objects that were not present in the training dataset. This network architecture can be used in building robots that can interact naturally with humans and learn about new objects over time.





\section*{ACKNOWLEDGMENT}

This project was supported by the Robert Bosch Center for Cyber-Physical Systems.


\bibliographystyle{IEEEtran}
\bibliography{IEEEabrv,mybibfile} 




\end{document}